\documentclass[runningheads]{llncs}

\usepackage{graphicx}
\usepackage{multirow}
\usepackage[mathletters]{ucs}
\usepackage[utf8x]{inputenc}
\usepackage{adjustbox}
\usepackage{hyperref}
\usepackage{csquotes}
\usepackage{xcolor,colortbl}
\usepackage{caption}
\usepackage{etoolbox}
\usepackage{multirow}

\AtBeginEnvironment{quote}{\singlespace\vspace{-\topsep}\small}
\AtEndEnvironment{quote}{\vspace{-\topsep}\endsinglespace}
\definecolor{Gray}{gray}{0.9}

\begin{document}
\title{TexTrolls: Identifying Russian Trolls on Twitter from a Textual Perspective}
%
%
\author{Bilal Ghanem\inst{1} \and
Davide Buscaldi\inst{2} \and
Paolo Rosso\inst{1} }
\authorrunning{Ghanem et al.}

\authorrunning{Anonymous et al.}

\institute{
PRHLT Research Center, Universitat Polit\`ecnica de Val\`encia, Spain \\
\email{\{bigha@doctor, prosso@dsic\}.upv.es} \and
CNRS / Laboratoire LIPN, Universit\'e Paris 13, France \\
\email{buscaldi@lipn.univ-paris13.fr}
}

\maketitle              
\begin{abstract}
The online new emerging suspicious users, that usually are called trolls, are one of the main sources of hate, fake, and deceptive online messages. Some agendas are utilizing these harmful users to spread incitement tweets, and as a consequence, the audience get deceived. The challenge in detecting such accounts is that they conceal their identities which make them disguised in social media, adding more difficulty to identify them using just their social network information. Therefore, in this paper, we propose a text-based approach to detect the online trolls such as those that were discovered during the US 2016 presidential elections. Our approach is mainly based on textual features which utilize thematic information, and profiling features to identify the accounts from their way of writing tweets. We deduced the thematic information in a unsupervised way and we show that coupling them with the textual features enhanced the performance of the proposed model. In addition, we find that the proposed profiling features perform the best comparing to the textual features.

\end{abstract}

\section{Introduction}
Recent years have seen a large increase in the amount of disinformation and fake news spread on social media. False information was used to spread fear and anger among people, which in turn, provoked crimes in some countries. The US in the recent years experienced many similar cases during the presidential elections, such as the one commonly known as ``Pizzagate" \footnote{\url{https://www.rollingstone.com/politics/politics-news/anatomy-of-a-fake-news-scandal-125877/}}. Later on, Twitter declared that they had detected a suspicious campaign originated in Russia by an organization named Internet Research Agency (IRA), and targeted the US to affect the results of the 2016 presidential elections\footnote{\url{https://blog.twitter.com/official/en_us/topics/company/2018/2016-election-update.html}}. The desired goals behind these accounts are to spread fake and hateful news to further polarize the public opinion. Such attempts are not limited to Twitter, since Facebook announced in mid-2019 that they detected a similar attempt originating from UAE, Egypt and Saudi Arabia and targeting other countries such as Qatar, Palestine, Lebanon and Jordan\footnote{\url{https://newsroom.fb.com/news/2019/08/cib-uae-egypt-saudi-arabia/}}. This attempt used Facebook pages, groups, and user accounts with fake identities to spread fake news supporting their ideological agendas. The automatic detection of such attempts is very challenging, since the true identity of these suspicious accounts is hidden by imitating the profiles of real persons from the targeted audience; in addition, sometimes they publish their suspicious idea in a vague way through their tweets' messages.

A previous work \cite{im2019still} showed that such suspicious accounts are not bots in a strict sense and they argue that they could be considered as ``software-assisted human workers". According to \cite{clark2016sifting}, the online suspicious accounts can be categorized into 3 main types: Robots, Cyborgs, and Human Spammers. We consider IRA accounts as another new emerging type called trolls, which is similar to Cyborgs except that the former focuses on targeting communities instead of individuals\footnote{ \url{https://itstillworks.com/difference-between-troll-cyberbully-5054.html}}. 

In this work, we identify online trolls in Twitter, namely IRA trolls, from a textual perspective. We study the effect of a set of text-based features and we propose a machine learning model to detect them\footnote{Data and model are available at: \url{https://www.github.com/anonymous}}. We aim to answer three research questions: RQ1. \textit{Does the thematic information improve the detection performance?}, RQ2. \textit{Can we detect IRA trolls from only a textual perspective?} and RQ3. \textit{How IRA campaign utilized the emotions to affect the public opinions?} 

The rest of the paper is structured as follows. In the following section, we present an overview on the literature work on IRA trolls. In Section \ref{lab:data}, we describe how the used dataset was compiled. Section \ref{lab:features} describes our proposed features for our approach. The experiments, results, and analyses are presented in Section \ref{lab:results}. Finally, we draw some conclusions and discuss possible future work on IRA trolls.

\section{Related Work on IRA Trolls}
\label{lab:related_Works}

After the 2016 US elections, Twitter has detected a suspicious attempt by a large set of accounts to influence the results of the elections. Due to this event, an emerging research works about the Russian troll accounts started to appear \cite{boyd2018characterizing,zannettou2019disinformation,im2019still,gorrell2019partisanship,badawy2019falls}.

The research works studied IRA trolls from several perspectives. The work in \cite{gorrell2019partisanship} studied the links' domains that were mentioned by IRA trolls and how much they overlap with other links used in tweets related to "Brexit". In addition, they compare "Left" and "Right" ideological trolls in terms of the number of re-tweets they received, number of followers, etc, and the online propaganda strategies they used. The authors in \cite{boyd2018characterizing} analyzed IRA campaign in both Twitter and Facebook, and they focus on the evolution of IRA paid advertisements on Facebook before and after the US presidential elections from a thematic perspective. 

The analysis work on IRA trolls not limited only to the tweets content, but it also considered the profile description, screen name, application client, geo-location, timezone, and number of links used per each media domain \cite{zannettou2019disinformation}. There is a probability that Twitter has missed some IRA accounts that maybe were less active than the others. Based on this hypothesis, the work in \cite{im2019still} built a machine learning model based on profile, language distribution, and stop-words usage features to detect IRA trolls in a newly sampled data from Twitter. Other works tried to model IRA campaign not only by focusing on the trolls accounts, but also by examining who interacted with the trolls by sharing their contents \cite{badawy2018analyzing}. Similarly, the work \cite{badawy2019falls} proposed a model that made use of the political ideologies of users, bot likelihood, and activity-related account metadata to predict users who spread the trolls’ contents.

\section{Data}
\label{lab:data}
To model the identification process of the Russian trolls, we considered a large dataset of both regular users (legitimate accounts) and IRA troll accounts. Following we describe the dataset. In Table \ref{dataset} we summarizes its statistics.

\begin{table}
\centering
\caption{Description of the dataset.}
\label{dataset}
\begin{tabular}{|l|c|c|c|}
\hline
\rowcolor{Gray}
{} &  Russian Trolls & Regular Accounts & Total \\
\hline 
Total \# of Accounts & { 2,023 } & { 94,643 } & { 96,666} \\
\hline
Total \# of Tweets & { 1,840,679 } & { 1,945,505 } & { 3,786,184 } \\
\hline
\end{tabular}
\end{table}

\subsection{Russian Trolls (IRA)}
We used the IRA dataset\footnote{\url{https://about.twitter.com/en_us/values/elections-integrity.html}} that was released by Twitter after identifying the Russian trolls. The original dataset contains $3,841$ accounts, but we use a lower number of accounts and tweets after filtering them. We focus on accounts that use English as main language. In fact, our goal is to detect Russian accounts that mimic a regular US user. Then, we remove from these accounts non-English tweets, and maintain only tweets that were tweeted originally by them. Our final IRA accounts list contains 2,023 accounts.

\subsection{Regular Accounts}
To contrast IRA behaviour, we sampled a large set of accounts to represent the ordinary behaviour of accounts from US. We collected a random sample of users that they post at least 5 tweets between 1$^{st}$ of August and 31 of December, 2016 (focusing on the US 2016 debates: first, second, third and vice president debates and the election day) by querying Twitter API hashtags related to the elections and its parties (e.g \#trump, \#clinton, \#election, \#debate, \#vote, etc.). In addition, we selected the accounts that have location within US and use English as language of the Twitter interface.
We focus on users during the presidential debates and elections dates because we suppose that the peak of trolls efforts concentrated during this period.

The final dataset is totally imbalanced (2\% for IRA trolls and 98\% for the regular users). This class imbalance situation represent a real scenario.
From Table \ref{dataset}, we can notice that the number of total tweets of the IRA trolls is similar to the one obtained from the regular users. This is due to the fact that IRA trolls were posting a lot of tweets before and during the elections in an attempt to try to make their messages reach the largest possible audience.

\section{Textual Representation}
\label{lab:features}

In order to identify IRA trolls, we use a rich set of textual features. With this set of features we aim to model the tweets of the accounts from several perspectives.

\subsection{Thematic Information} 
Previous works \cite{NG18} have investigated IRA campaign efforts on Facebook, and they found that IRA pages have posted more than $\sim$80K posts focused on division issues in US. Later on, the work in \cite{boyd2018characterizing} has analyzed Facebook advertised posts by IRA and they specified the main themes that these advertisements discussed. Given the results of the previous works, we applied a topic modeling technique on our dataset to extract its main themes. We aim to detect IRA trolls by identifying their suspicious ideological changes across a set of themes.

\begin{figure}[!htb]
   \begin{minipage}{0.49\textwidth}
    \centering
    \includegraphics[width=6cm]{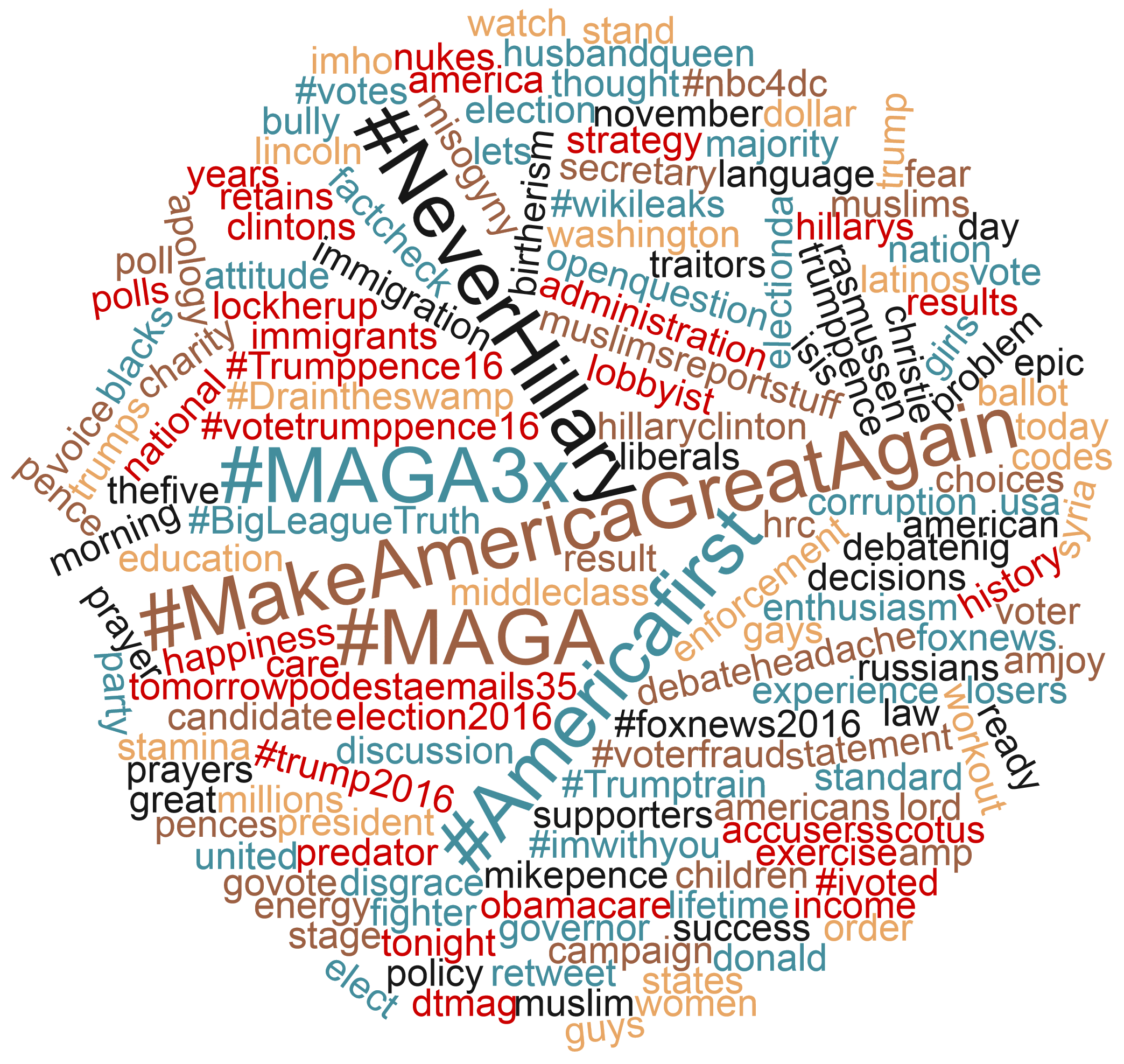}
    \captionsetup{labelformat=empty}
    (a)
   \end{minipage}\hfill
   \begin{minipage}{0.49\textwidth}
    \centering
    \includegraphics[width=6cm]{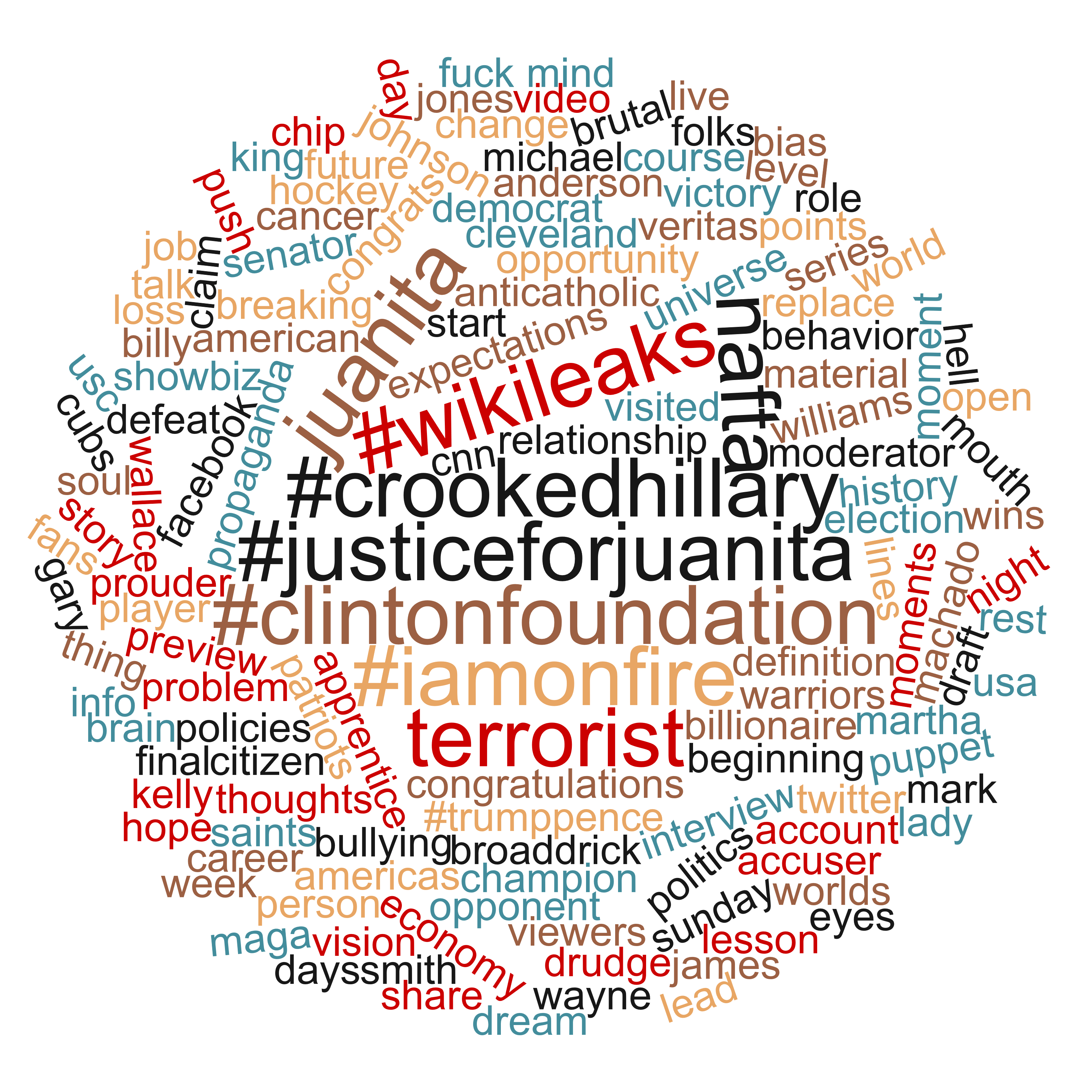}
    \captionsetup{labelformat=empty}
    (b)
    \end{minipage}
    \caption{(a) \textit{Supporting Trump} and (b) \textit{Attacking Hillary} themes words clouds.}
    \label{fig:wordclouds}
\end{figure}

Given our dataset, we applied Latent Dirichlet Allocation (LDA) topic modeling algorithm \cite{blei2003latent} on the tweets after a prepossessing step where we maintained only nouns and proper nouns. In addition, we removed special characters (except HASH "\#" sign for the hashtags) and lowercase the final tweet. To ensure the quality of the themes, we removed the hashtags we used in the collecting process where they may bias the modeling algorithm. We tested multiple number of themes and we chose seven of them. We manually observed the content of these themes to label them. The extracted themes are: \textit{Police shootings, Islam and War, Supporting Trump, Black People, Civil Rights, Attacking Hillary,} and \textit{Crimes}. In some themes, like \textit{Supporting Trump} and \textit{Attacking Hillary}, we found contradicted opinions, in favor and against the main themes, but we chose the final stance based on the most representative hashtags and words in each of them (see Figure \ref{fig:wordclouds}). Also, the themes \textit{Police Shooting} and \textit{Crimes} are similar, but we found that some words such as: \textit{police, officers, cops, shooting, gun, shot, etc.} are the most discriminative between these two themes. In addition, we found that the \textit{Crimes} theme focuses more on raping crimes against children and women. Our resulted themes are generally consistent with the ones obtained from the Facebook advertised posts in \cite{boyd2018characterizing}, and this emphasizes that IRA efforts organized in a similar manner in both social media platforms.

Based on our thematic information, we model the users textual features w.r.t. each of these themes. In other words, we model a set of textual features independently for each of the former themes to capture the emotional, stance, and others changes in the users tweets. 

For the theme-based features, we use the following features that we believe that they change based on the themes:
\begin{itemize}
    \item \textbf{Emotions}: Since the results of the previous works \cite{boyd2018characterizing,NG18} showed that IRA efforts engineered to seed discord among individuals in US, we use emotions features to detect their emotional attempts to manipulate the public opinions (e.g. fear spreading behavior). For that, we use the NRC emotions lexicon \cite{mohammad2010emotions} that contains $\sim$14K words labeled using the eight Plutchik's emotions. 
    
    \item \textbf{Sentiment}: We extract the sentiment of the tweets from NRC \cite{mohammad2010emotions}, \textit{positive} and \textit{negative}.
    
    \item \textbf{Bad \& Sexual Cues}: During the manual analysis of a sample from IRA tweets, we found that some users use bad slang word to mimic the language of a US citizen. Thus, we model the presence of such words using a list of bad and sexual words from \cite{frenda2018exploration}.
    
    \item \textbf{Stance Cues}: Stance detection has been studied in different contexts to detect the stance of a tweet reply with respect to a main tweet/thread \cite{mohammad2016semeval}. Using this feature, we aim to detect the stance of the users regarding the different topics we extracted. To model the stance we use a set of stance lexicons employed in previous works \cite{bahuleyan2017uwaterloo,ghanem2019upv}. Concretely, we focus on the following categories: \textit{belief, denial, doubt, fake, knowledge, negation, question,} and \textit{report}.
    
    \item \textbf{Bias Cues}: We rely on a set of lexicons to capture the bias in text. We model the presence of the words in one of the following cues categories: \textit{assertives verbs} \cite{hooper1974assertive}, \textit{bias} \cite{recasens2013linguistic}, \textit{factive verbs} \cite{kiparsky1968fact}, \textit{implicative verbs} \cite{karttunen1971implicative}, \textit{hedges} \cite{hyland2018metadiscourse}, \textit{report verbs} \cite{recasens2013linguistic}. A previous work has used these bias cues to identify bias in suspicious news posts in Twitter \cite{volkova2018predicting}.
    
    \item \textbf{LIWC}: We use a set of linguistic categories from the LIWC linguistic dictionary \cite{tausczik2010psychological}. The used categories are: \textit{pronoun, anx, cogmech, insight, cause, discrep, tentat, certain, inhib, incl}\footnote{Total pronouns, Anxiety, Cognitive processes, Insight, Causation, Discrepancy, Tentative, Certainty, Inhibition, and Inclusive respectively.}.
    
    \item \textbf{Morality}: Cues based on the morality foundation theory \cite{graham2009liberals} where words labeled in one of a set of categories: \textit{care, harm, fairness, cheating, loyalty, betrayal, authority, subversion, sanctity,} and \textit{degradation}.
        
\end{itemize}

Given $V_i$ as the concatenation of the previous features vectors of a tweet$_i$, we represent each user's tweets by considering the average and standard deviation of her tweets' $V_{1,2,..N}$ in each theme $j$ independently and we concatenate them. Mathematically, a user $x$ final feature vector is defined as follows:

\begin{equation}
user_x = \bigodot_{j=1}^{T}
\left[
\frac{\sum^{N_j}_{i=1}V_{ij}}{{N_j}}\odot\sqrt{ \frac{\sum^{N_j}_{i=1}(V_{ij}-\overline{{V_j}})}{{N_j}} } 
\right]
\end{equation}

where given the j$^{th}$ theme, $N_j$ is the total number of tweets of the user, $V_{ij}$ is the i$^{th}$ tweet feature vector, $\overline{V_j}$ is the mean of the tweets' feature vectors. With this representation we aim at capturing the "Flip-Flop" behavior of IRA trolls among the themes (see Section \ref{analysis}).

\subsection{Profiling IRA Accounts}
As Twitter declared, although the IRA campaign was originated in Russia, it has been found that IRA trolls concealed their identity by tweeting in English. Furthermore, for any possibility of unmasking their identity, the majority of IRA trolls changed their location to other countries and the language of the Twitter interface they use. Thus, we propose the following features to identify these users using only their tweets text:   

\begin{itemize}
    \item \textbf{Native Language Identification (NLI)}: This feature was inspired by earlier works on identifying native language of essays writers \cite{malmasi2017report}. We aim to detect IRA trolls by identifying their way of writing English tweets. As shown in \cite{volkova2018predicting}, English tweets generated by non-English speakers have a different syntactic pattern . Thus, we use state-of-the-art NLI features to detect this unique pattern \cite{cimino2017stacked,markov2017cic,goutte2017exploring}; the feature set consists of bag of stopwords, Part-of-speech tags (POS), and syntactic dependency relations (DEPREL). We extract the POS and the DEPREL information using spaCy\footnote{\url{https://spacy.io/models}}, an off-the-shelf POS tagger\footnote{We tested TweetNLP \cite{owoputi2013improved}, the state-of-the-art POS tagger on tweets, but spaCy produced higher overall results.}. We clean the tweets from the special characters and maintained dots, commas, and first-letter capitalization of words. We use regular expressions to convert a sequence of dots to a single dot, and similarly for sequence of characters.
    
    \item \textbf{Stylistic}: We extract a set of stylistic features following previous works in the authorship attribution domain \cite{zheng2006framework,bhargava2013stylometric,sultana2017authorship}, such as: the count of special characters, consecutive characters and letters\footnote{We considered 2 or more consecutive characters, and 3 or more consecutive letters.}, URLs, hashtags, users' mentions. In addition, we extract the uppercase ratio and the tweet length.
\end{itemize}{}
Similar to the feature representation of the theme-based features, we represent each user's tweets by considering the average and standard deviation of her tweets' $V_{1,2,..N}$, given $V_i$ as the concatenation of the previous two features vectors of a tweet$_i$. A user $x$ final feature vector is defined as follows:

\begin{equation}
user_x = 
\left[
\frac{\sum^N_{i=1}V_{i}}{N};\sqrt{ \frac{\sum^N_{i=1}(V_{i}-\overline{V})}{N} } 
\right]
\end{equation}

where $N$ is her total number of tweets, $V_i$ is the i$th$ tweet feature vector, $\overline{V}$ is the mean of her tweets feature vectors.

\section{Experiments and Analysis}
\label{lab:results}

\subsection{Experimental Setup}
We report precision, recall and F1 score. Given the substantial class imbalance in the dataset, we use the macro weighted version of the F1 metric.
We tested several classifiers and Logistic Regression showed the best F1$_{macro}$ value. We kept the default parameters values. We report results for 5-folds cross-validation.

\subsection{Baselines}
In order to evaluate our feature set, we use \textit{Random Selection}, \textit{Majority Class}, and \textit{bag-of-words} baselines. In the bag-of-words baseline, we aggregate all the tweets of a user into one document. A previous work \cite{boatwright2018troll} showed that IRA trolls were playing a hashtag game which is a popular word game played on Twitter, where users add a hashtag to their tweets and then answer an implied question \cite{haskell2015}. IRA trolls used this game in a similar way but focusing more on offending or attacking others; an example from IRA tweets: "\textit{\#OffendEveryoneIn4Words undocumented immigrants are ILLEGALS}". Thus, we use as a baseline \textit{Tweet2vec} \cite{dhingra2016tweet2vec} which is a a character-based Bidirectional Gated Recurrent neural network reads tweets and predicts their hashtags. We aim to assess if the tweets hashtags can help identifying the IRA tweets. The model reads the tweets in a form of character one-hot encodings and uses them for training with their hashtags as labels. To train the model, we use our collected dataset which consists of $\sim$3.7M tweets\footnote{We used the default parameters that were provided with the system code.}. To represent the tweets in this baseline, we use the decoded embedding produced by the model and we feed them to the Logistic Regression classifier.

IRA dataset provided by Twitter contains less information about the accounts details, and they limited to: profile description, account creation date, number of followers and followees, location, and account language. Therefore, as another baseline we use the number of followers and followees to assess their identification ability (we will mention them as \textit{Network Features} in the rest of the paper).

\subsection{Results}
Table \ref{tab1} presents the classification results showing the performance of each feature set independently. Generally, we can see that the thematic information improves the performance of the proposed features clearly (RQ1), and with the largest amount in the \textit{Emotions} features (see $-_{themes}$ and $+_{themes}$ columns). This result emphasizes the importance of the thematic information. Also, we see that the emotions performance increases with the largest amount considering F1$_{macro}$ value; this motivates us to analyze the emotions in IRA tweets (see the following section).

The result of the \textit{NLI} feature in the table is interesting; we are able to detect IRA trolls from their writing style with a F1$_{macro}$ value of 0.91. Considering the results in Table \ref{tab1}, we can notice that we are able to detect the IRA trolls effectively using only textual features (RQ2).

Finally, the baselines results show us that the \textit{Network} features do not perform well. A previous work \cite{zannettou2019disinformation} showed that IRA trolls tend to follow a lot of users, and nudging other users to follow them (e.g. by writing "follow me" in their profile description) to fuse their identity (account information) with the regular users. Finally, similar to the \textit{Network} features, the \textit{Tweet2vec} baseline performs poorly. This indicates that, although IRA trolls used the hashtag game extensively in their tweets, the \textit{Tweet2vec} baseline is not able to identify them.

\begin{table}
\centering
\caption{Classification results. We report the results of each feature set independently.}
\label{tab1}
\begin{tabular}{|l|c|c|c|c|}
\hline
\rowcolor{Gray}
{Method} & {Precision$_{macro}$} & {Recall$_{macro}$} & \multicolumn{2}{|c|}{F1$_{macro}$} \\
\hline
Random Selection & {0.5} & {0.5} & \multicolumn{2}{|c|}{0.35} \\
\hline
Majority Class & {0.49} & {0.5} & \multicolumn{2}{|c|}{0.5} \\
\hline
Tweet2vec & {0.49} & {0.5} & \multicolumn{2}{|c|}{0.5}\\
\hline
Network Features & {0.49} & {0.5} & \multicolumn{2}{|c|}{0.5} \\
\hline 
\rowcolor{Gray}
\multicolumn{1}{c}{} & \multicolumn{2}{c|}{Theme-based Features} & $_{\textbf{-}themes}$  & $_{+themes}$  \\
\hline
Emotions & {0.59} & {0.5} & {0.5} & {0.87} \\
\hline
Sentiment & {0.49} & {0.5} & {0.5} & {0.64} \\
\hline 
Bad \& Sexual Cues & {0.49} & {0.5} & {0.5} & {0.79}  \\
\hline
Stance cues & {0.91} & {0.53} & {0.56} & {0.86}  \\
\hline
Bias Cues & {0.73} & {0.51} & {0.51} & {0.87} \\
\hline
LIWC & {0.69} & {0.51} & {0.51} & {0.85} \\
\hline
Morality & {0.90} & {0.62} & {0.68} & {0.86} \\
\hline
\rowcolor{Gray}
\multicolumn{5}{|c|}{Profiling Features} \\
\hline
Stylistic & {0.92} & {0.84} & \multicolumn{2}{|c|}{0.88} \\
\hline
NLI & {0.96} & {0.88} & \multicolumn{2}{|c|}{0.91} \\
\hline
\hline
\textbf{All Features} & {0.96} & {0.93} & \multicolumn{2}{|c|}{0.94} \\
\hline
\end{tabular}
\end{table}

\subsection{Analysis}
\label{analysis}
\begin{figure}[!htb] 
  \centering
  \includegraphics[width=10cm]{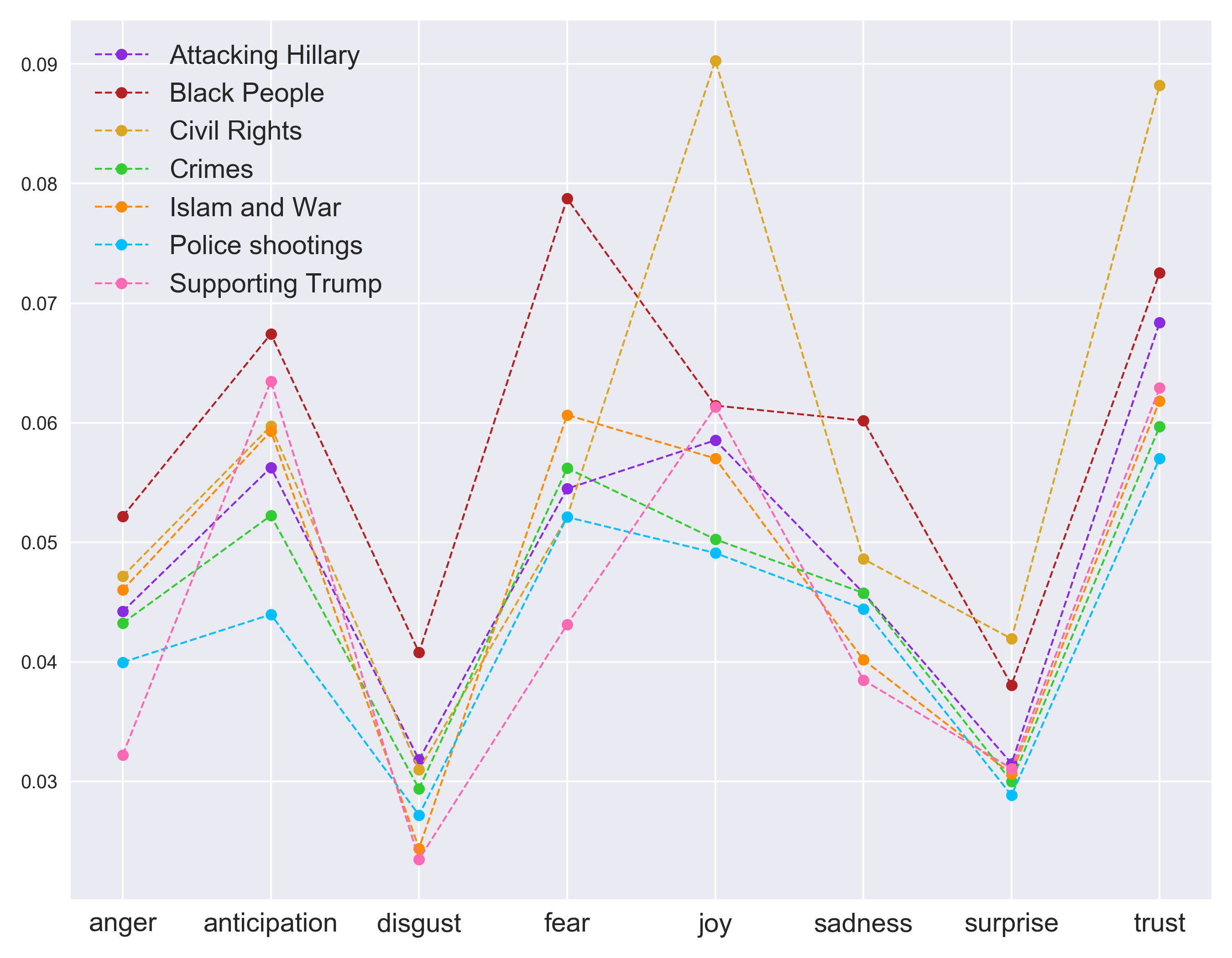}
  \caption{Emotional analyzing of IRA trolls from thematic perspective.}
  \label{fig:emotions}
\end{figure}

Given that the \textit{Emotions} features boosted the F1$_{macro}$ with the highest value comparing to the other theme-based features, in Figure \ref{fig:emotions} we analyze IRA trolls from emotional perspective to answer RQ3.
The analysis shows that the themes that were used to attack immigrants (\textit{Black People} and \textit{Islam and War}) have the \textit{fear} emotion in their top two emotions. While on the other hand, a theme like \textit{Supporting Trump} has a less amount of \textit{fear} emotion, and the \textit{joy} emotion among the top emotions.

\noindent \textit{Why do the thematic information help? The Flip-Flop behavior.} As an example, let's considering the \textit{fear} and \textit{joy} emotions in Figure \ref{fig:emotions}. We can notice that all the themes that used to nudge the division issues have a decreasing dashed line, where others such as \textit{Supporting Trump} theme has an extremely increasing dashed line. Therefore, we manually analyzed the tweets of some IRA accounts and we found this observation clear, as an example from user $x$:

\begin{displayquote}
\textit{Islam and War}: \textbf{(A)} @RickMad: Questions are a joke, a Muslim asks how SHE will be protected from Islamaphobia! Gmaffb! How will WE be protected from terrori…
\end{displayquote}

\begin{displayquote}
\textit{Supporting Trump}: \textbf{(B)} @realDonaldTrump: That was really exciting. Made all of my points. MAKE AMERICA GREAT AGAIN!
\end{displayquote}
 
Figure \ref{fig:flipflop} shows the flipping behaviour for user $x$ by extracting the mean value of the \textit{fear} and \textit{joy} emotions. The smaller difference between the \textit{fear} and \textit{joy} emotions in the \textit{Islam and War} theme for this user is due to the ironic way of tweeting for the user (e.g. the beginning of tweet A: "Questions are a joke"). Even though, the \textit{fear} emotion is still superior to the \textit{joy}. 
We notice a similar pattern in some of the regular users, although much more evident among IRA trolls. 

\begin{figure}[!h] 
  \centering
  \includegraphics[width=6.5cm]{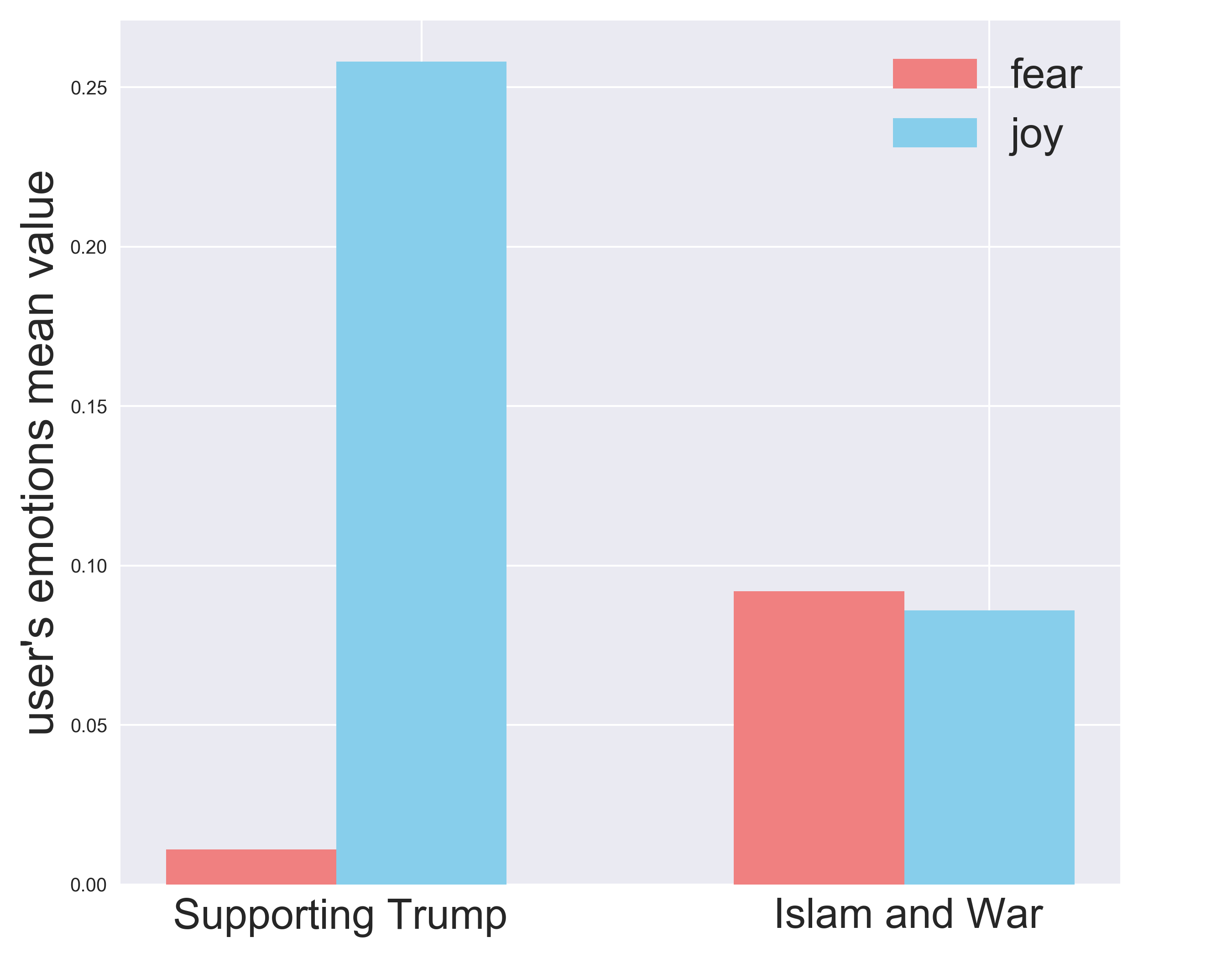}
  \caption{Flipping emotions between themes by user $x$ (an IRA troll).}
  \label{fig:flipflop}
\end{figure}

To understand more the \textit{NLI} features performance, given their high performance comparing to the other features, we extract the top important tokens for each of the \textit{NLI} feature subsets (see Figure \ref{fig:top10}). Some of the obtained results confirmed what was found previously. For instance, the authors in \cite{volkova2018predicting} found that Russians write English tweets with more prepositions comparing to native speakers of other languages (e.g. as, about, because in (c) Stop-words and RP\footnote{RP stands for "adverb, particle" in the POS tag set.} in (a) POS in Figure \ref{fig:top10}). Further research must be conducted to investigate in depth the rest of the results.

\begin{figure}[!htb]
\centering
   \begin{minipage}{0.33\textwidth}
   \addtocounter{figure}{-1}
    \centering
    \includegraphics[width=3cm]{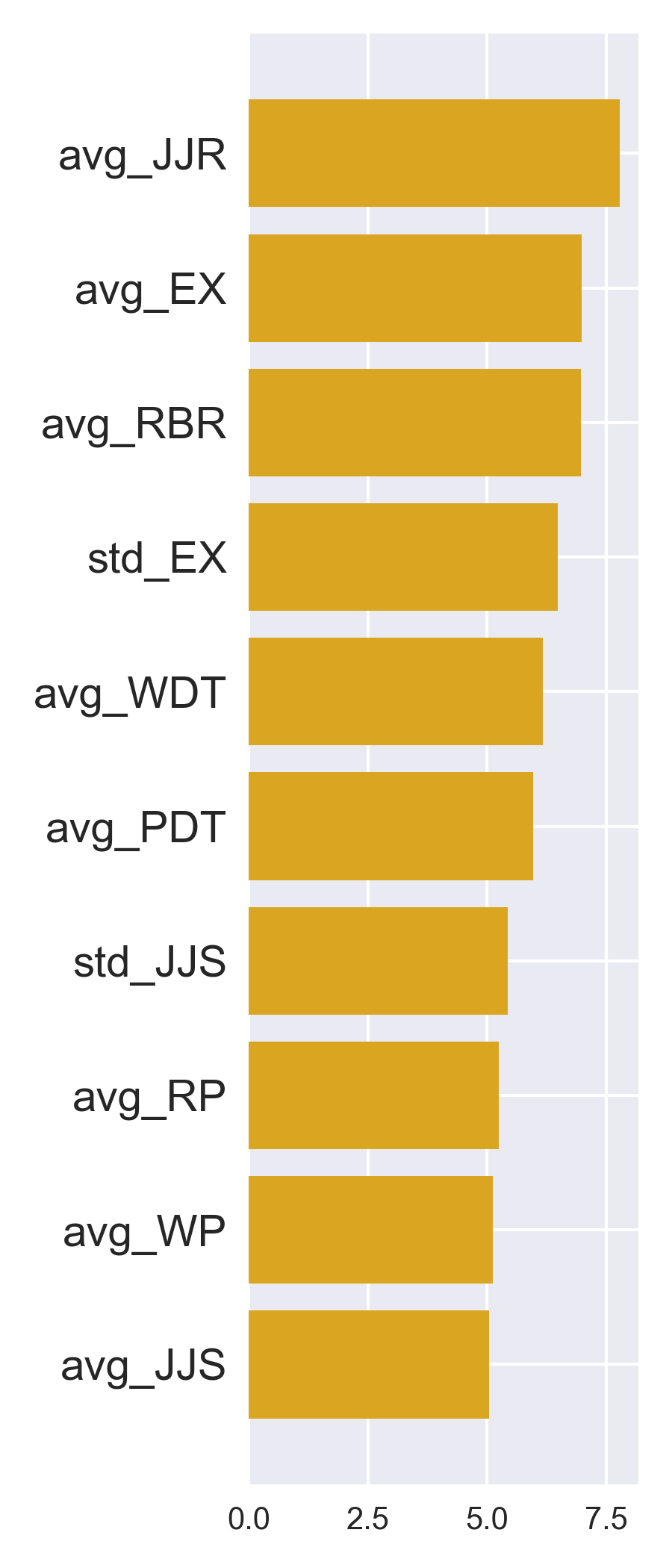}
    \captionsetup{labelformat=empty}
    \caption{(a) POS}
    \end{minipage}\hfill
   \begin{minipage}{0.33\textwidth}
    \addtocounter{figure}{-1}
    \centering
    \includegraphics[width=3cm]{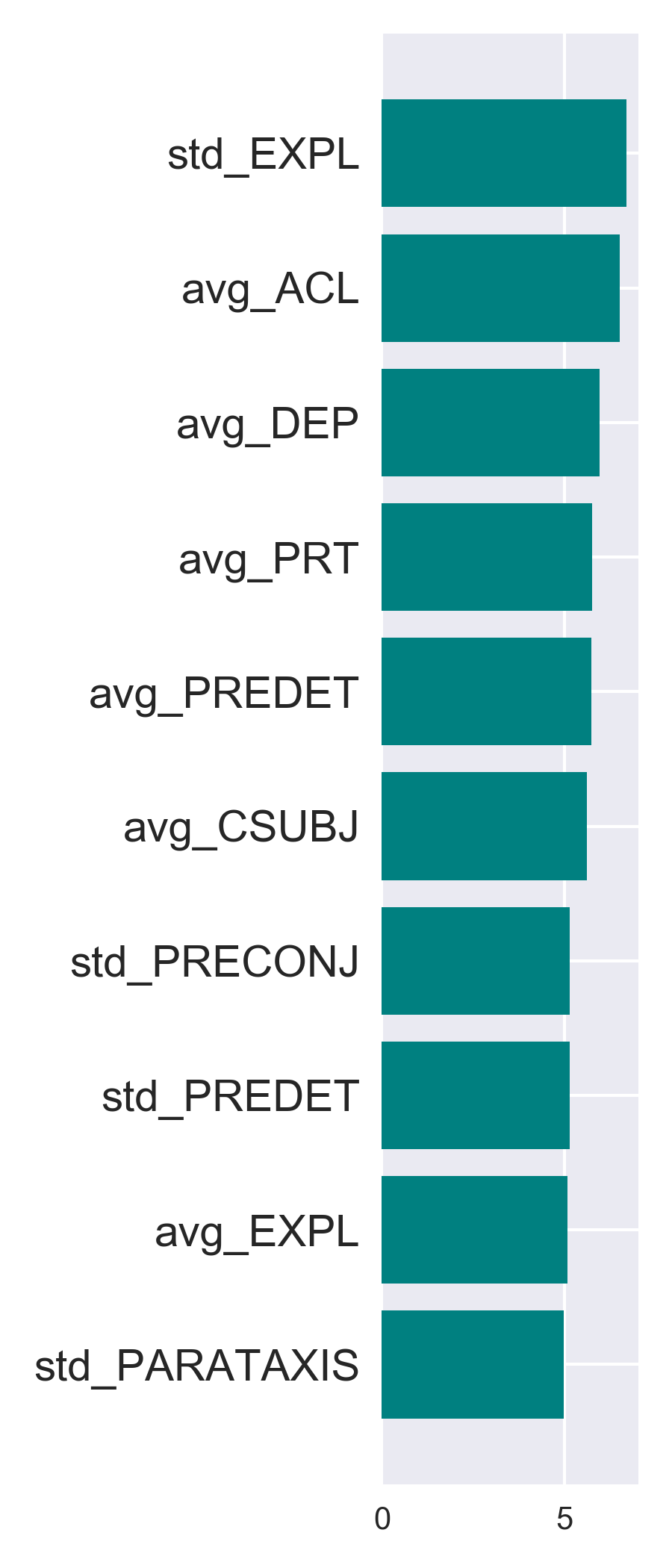}
    \captionsetup{labelformat=empty}
    \caption{(b) DEPREL}
    \end{minipage}
    \begin{minipage}{0.33\textwidth}
    \addtocounter{figure}{-1}
    \centering
    \includegraphics[width=3cm]{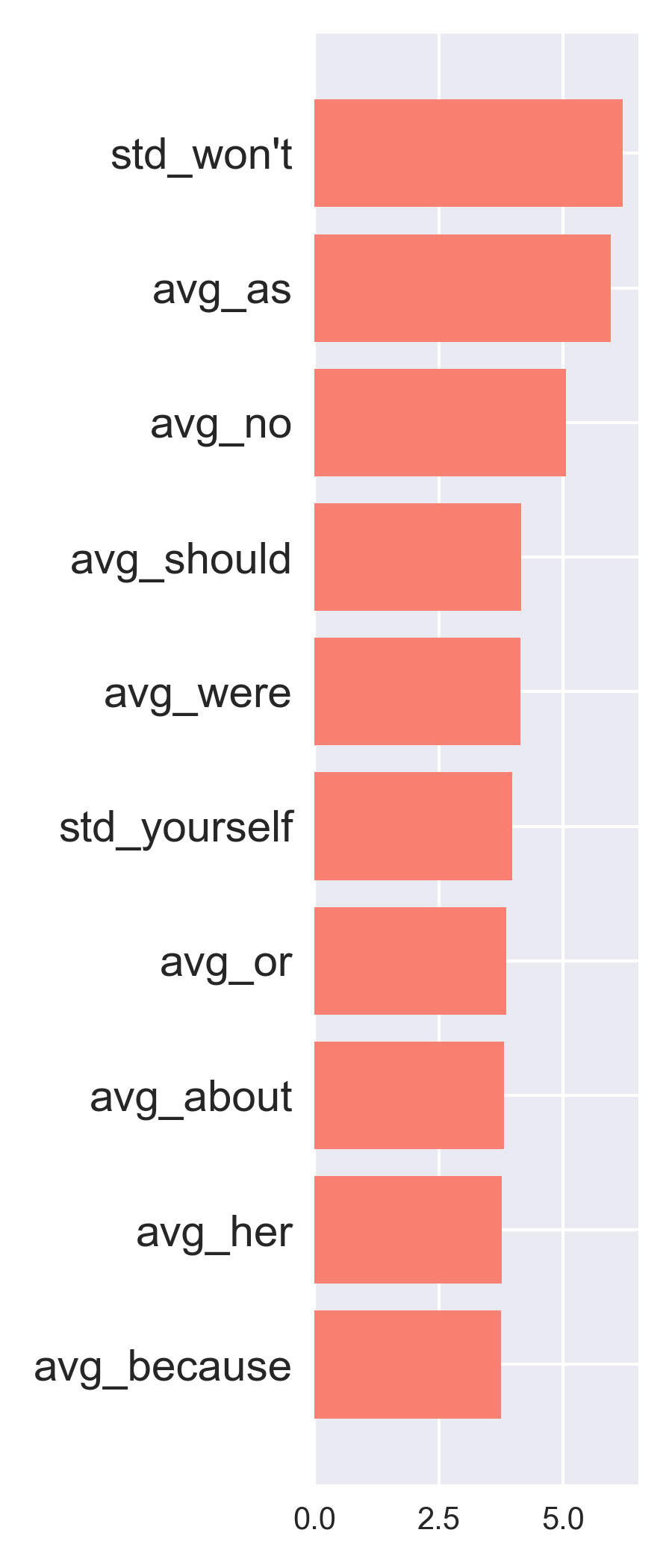}
    \captionsetup{labelformat=empty}
    \caption{(c) Stop-words}
    \end{minipage}\hfill
    \caption{The top 10 important tokens in each of the NLI features.}
    \label{fig:top10}
\end{figure}

\noindent\textbf{Linguistic Analysis.} We measure statistically significant differences in the cues markers of Morality, LIWC, Bias and Subjectivity, Stance, and Bad and Sexual words across IRA trolls and regular users. These findings presented in Table \ref{tab:linguistic} allows for a deeper understanding of IRA trolls.

\begin{table}[h]
\centering
\caption{Linguistic analysis of Morality, LIWC, Bias and Subjectivity, Stance, and Bad and Sexual cues shown as the \textbf{percentage of averaged value of tweets with one or more cues} across IRA trolls (X) and regular users (Y) in a shape of X(arrows)Y. We report only significant differences: p-value ≤ 0.001↑↑↑, ≤ 0.01↑↑, ≤ 0.05↑ estimated using the Mann-Whitney U test. The tweets average value is the mean value across the themes.}
\label{tab:linguistic}
\begin{tabular}{|c|c|c|c|c|c|}
\hline
\rowcolor{Gray}
\multicolumn{2}{|c|}{Morality} & \multicolumn{2}{|c|}{LIWC} & \multicolumn{2}{|c|}{Bias language} \\
\hline
category & $P_{value}$ & category & $P_{value}$ & category & $P_{value}$ \\
\hline   
care       & 1.39$↑↑↑$0.74 & pronoun & 53.34$↑$47.59      & assertive   & 6.53$↓↓↓$7.05      \\
\hline
harm       & 2.39$↑↑↑$1.61 & anx     & 1.91$↑↑↑$0.98      & bias        & ×                \\
\hline
fairness   & 0.64$↓↓↓$0.84 & cogmech & ×                & factive     & 5.58$↑↑↑$3.95      \\
\hline
cheating   & 0.06$↓↓$0.31  & insight & 12.16$↑↑↑$10.08    & hedge       & 10.06$↑↑↑$9.69     \\
\hline
loyalty    & 0.84$↓↓↓$1.26 & cause   & 10.76$↑↑↑$10.27    & implicative & 9.09$↑↑↑$6.37      \\
\hline
betrayal   & 0.13$↓↓↓$0.35 & discrep & 12.73$↑↑↑$11.07    & report      & 14.37$↓↓↓$18.89    \\
\hline
authority  & 1.59$↓↓↓$1.88 & tentat  & 13.95$↑↑↑$12.29    & strong subj & 54.15$↑↑↑$49.9     \\
\hline
subversion & 0.38$↑↑↑$0.33 & certain & 13.59$↑↑↑$10.69    & weak subj   & 50.33$↑$41.96      \\
\hline
Sanctity   & 0.44$↑↑↑$0.27 & inhib   & 4.19$↑↑↑$3.87      & -           & -                \\
\hline
degradation & 0.54$↑↑↑$0.49 & incl    & 20.69$↓↓$21.24     & -           & -               \\
\hline
\end{tabular}

\begin{tabular}{|c|c|c|c|}
\hline
\rowcolor{Gray}
\multicolumn{2}{|c|}{Stance} & \multicolumn{2}{|c|}{Bad and Sexual} \\
\hline
category & $P_{value}$ & category & $P_{value}$ \\
\hline   
belief      & 2.97$↑↑↑$2.49     & bad    & 5.45$↑↑↑$4.66 \\
\hline
denial      & 0.63$↑↑↑$0.57     & sexual & 3.54$↑↑↑$3.16 \\
\hline
doubt       & 1.34$↑↑↑$1.25     & -      & - \\
\hline
fake        & 0.49$↓↓↓$1.22     & -      & - \\
\hline
knowledge   & 0.75$↓↓↓$1.48     & -      & - \\
\hline
negation    & 11.46$↑↑↑$9.10    & -      & - \\
\hline
question    & 3.12$↑↑↑$2.44     & -      & - \\
\hline
report      & 2.86$↓↓↓$3.46     & -      & - \\
\hline
\end{tabular}
\end{table}

\noindent \textbf{False Positive Cases.} The proposed features showed to be effective in the classification process. We are interested in understanding the causes of misclassifying some of IRA trolls. Therefore, we manually investigated the false positive tweets and we found that there are three main reasons: 1) Some trolls were tweeting in a questioning way by asking about general issues; we examined their tweets but we did not find a clear ideological orientation or a suspicious behaviour in their tweets. 2) Some accounts were sharing traditional social media posts (e.g. "http://t.co/GGpZMvnEAj cat vs trashcan"); the majority of the false positive IRA trolls are categorized under this reason. In addition, these posts were given a false theme name; the tweet in the previous example assigned to \textit{Attacking Hillary} theme. 3) Lack of content. Some of the misclassified trolls mention only external links without a clear textual content. This kind of trolls needs a second step to investigate the content of the external links. Thus, we tried to read the content of these links but we found that the majority of them referred to deleted tweets. Probably this kind of accounts was used to "raise the voice" of other trolls, as well as, we argue that the three kinds of IRA trolls were used for "likes boosting".

\section{Conclusion}

In this work, we present a textual approach to detect social media trolls, namely IRA accounts. Due to the anonymity characteristic that social media provide to users, these kinds of suspicious behavioural accounts have started to appear. We built a new machine learning model based on theme-based and profiling features that in cross-validation evaluation achieved a F1$_{macro}$ value of 0.94. We applied a topic modeling algorithm to go behind the superficial textual information of the tweets. Our experiments showed that the extracted themes boosted the performance of the proposed model when coupled with other surface text features. In addition, we proposed \textit{NLI} features to identify IRA trolls from their writing style, which showed to be very effective. Finally, for a better understanding we analyzed the IRA accounts from emotional and linguistic perspectives.

Through the manually checking of IRA accounts, we noticed that frequently irony was employed. As a future work, it would be interesting to identify these accounts by integrating an irony detection module.


%
\bibliographystyle{splncs04}
\bibliography{references}
\end{document}